\documentclass[sigconf]{acmart}
\usepackage{caption}
\usepackage{subcaption}
\usepackage{multirow}
\usepackage{multicol}
\newcommand{\bftab}{\fontseries{b}\selectfont}
\usepackage{amsthm}
\usepackage{amsmath}
\usepackage{stmaryrd}
\usepackage{float}
\usepackage{stfloats}
\usepackage{color, colortbl}

\usepackage{hyperref}
\newtheorem*{prop}{Proposition}

\copyrightyear{2022}
\acmYear{2022}
\setcopyright{acmcopyright}
\acmConference[KDD '22] {Proceedings of the 28th ACM SIGKDD Conference on Knowledge Discovery and Data Mining}{August 14--18, 2022}{Washington, DC, USA.}
\acmBooktitle{Proceedings of the 28th ACM SIGKDD Conference on Knowledge Discovery and Data Mining (KDD '22), August 14--18, 2022, Washington, DC, USA}
\acmPrice{15.00}
\acmISBN{978-1-4503-9385-0/22/08}
\acmDOI{10.1145/3534678.3539457}

\begin{document}

\title{Learning on Graphs with Out-of-Distribution Nodes}

\author{Yu Song}
\email{songyu132@outlook.com}
\orcid{0000-0002-8940-2561}
\affiliation{%
  \institution{Westlake University}
  \city{Hangzhou}
  \country{China}
}
\author{Donglin Wang}
\authornote{Corresponding author.}
\email{wangdonglin@westlake.edu.cn}
\orcid{0000-0002-8188-3735}
\affiliation{%
  \institution{Westlake University}
  \city{Hangzhou}
  \country{China}
}
\begin{abstract}
Graph Neural Networks (GNNs) are state-of-the-art models for performing prediction tasks on graphs.
While existing GNNs have shown great performance on various tasks related to graphs, little attention has been paid to the scenario where out-of-distribution (OOD) nodes exist in the graph during training and inference. 
Borrowing the concept from CV and NLP, we define OOD nodes as nodes with labels unseen from the training set. Since a lot of networks are automatically constructed by programs, real-world graphs are often noisy and may contain nodes from unknown distributions. In this work, we define the problem of \textit{graph learning with out-of-distribution nodes}. Specifically, we aim to accomplish two tasks: 1) detect nodes which do not belong to the known distribution and 2) classify the remaining nodes to be one of the known classes. We demonstrate that the connection patterns in graphs are informative for outlier detection, and propose Out-of-Distribution Graph Attention Network (OODGAT), a novel GNN model which explicitly models the interaction between different kinds of nodes and separate inliers from outliers during feature propagation. Extensive experiments show that OODGAT outperforms existing outlier detection methods by a large margin, while being better or comparable in terms of in-distribution classification.
\end{abstract}




\maketitle

\section{Introduction}
Graphs neural networks (GNNs) have become the \textit{de facto} tool for performing prediction tasks on graphs. Among various applications, one of the most important tasks of GNNs is semi-supervised node classification (SSNC) \cite{kipf2016semi}. In SSNC, GNNs aggregate information from adjacent nodes and generate representations that are smooth within neighborhoods, alleviating the difficulty of classification. 

In recent years, many studies have begun to consider graph learning tasks in realistic settings, such as graphs with label noise \cite{nrgnn2021}, low labeling rates \cite{few2020} and distribution shifts \cite{generalization2021, size2021}. However, very few work has considered the scenario where out-of-distribution (OOD) nodes exist in the graph on which one performs SSNC. By using the term 'OOD', we borrow the notion from CV and NLP, which means samples with labels not seen in the training set. In the graph domain, this can be quite common as graphs are usually constructed in an incremental way where new nodes are added due to the connectivity with existing ones, and for most cases there is no guarantee that all nodes must connect to others from the same distribution. For example, we want to classify papers in a citation network into AI-related topics, e.g., deep learning, reinforcement learning and optimization methods. The network is obtained using a web crawler which adopts a breadth first search (BFS) strategy and keeps exploring papers referencing existing ones for a number of iterations. When searching stops, the resulting network is not guaranteed to contain nodes only from the known categories, as it is common for a scientific paper to refer to articles in less relevant research areas, for example, an AI paper might cite papers in neuroscience and mathematics. In real-world networks, the proportion of nodes from irrelevant categories may even be higher than those from the classes of interest. \textit{Given a noisy graph as such, our task is to predict the label for nodes which correspond to one of the known classes, and identify nodes that do not belong to any of them.} 

\begin{figure}
    \centering
    \includegraphics[scale=0.40]{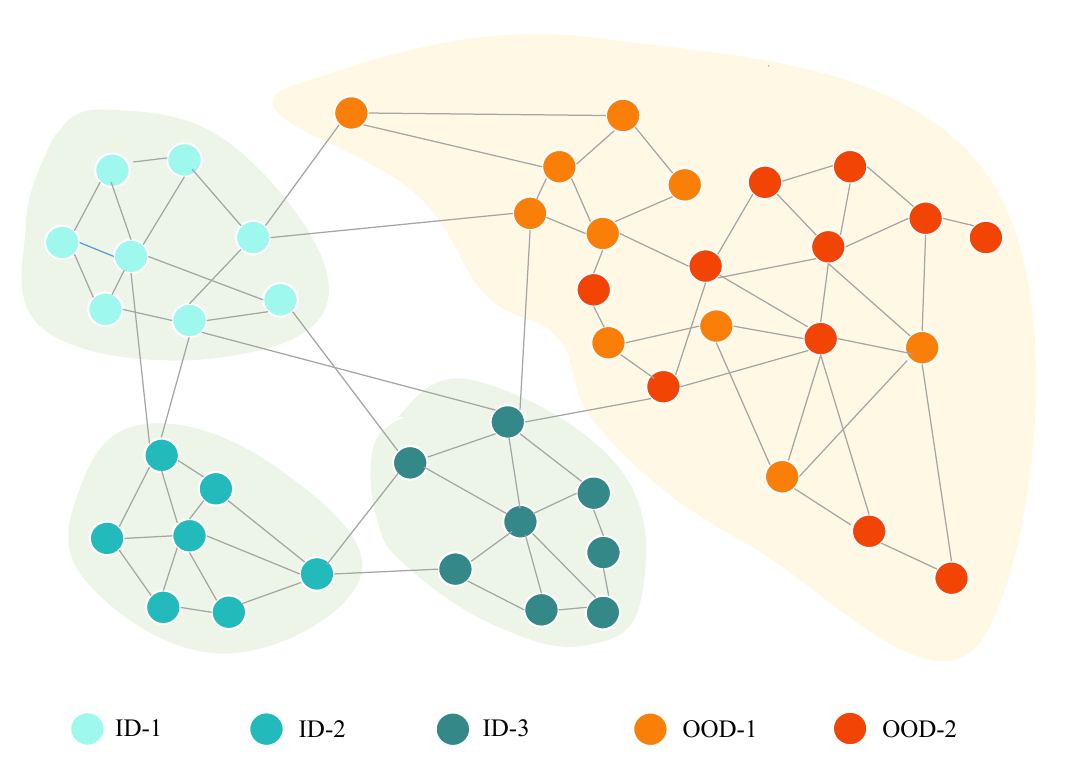}
    \caption{An illustration of graph learning with out-of-distribution nodes. In this setting, we aim to accomplish two tasks: 1) separate ID nodes from OOD nodes and 2) classify ID nodes correctly. Colors of nodes indicate their labels, and the shaded areas represent a possible set of  decision boundaries. Note that the connections exist not only within ID nodes, but also within OOD nodes, and in between. Unlike traditional anomaly detection which assumes a small percentage of anomalies, in the graph domain, the OOD part may contain nodes that are comparable in size to the ID part.}
    \label{fig:illustration}
\end{figure}

In CV and NLP, OOD detection has been a hot research area with a long history. \cite{msp2016} demonstrates that neural networks tend to assign higher softmax probabilities to in-distribution (ID) samples than to out-of-distribution (OOD) ones, and proposes to use the maximum softmax probability (MSP) produced by the neural network as the score for OOD detection. Other approaches attempt to improve detection performance by modifying the model structure \cite{mcd2019, eloc2018}, employing customized uncertainty measures \cite{m2018} or exploiting labeled outliers \cite{oe2018}.

Different from the above methods which only focus on identifying OOD samples at inference time, the presence of OOD nodes in graphs makes the task more challenging. First, in traditional settings of CV and NLP, outliers only occur in the test set, while in the graph domain one is usually given the entire graph for training which consists of both inliers and outliers, transferring the problem from detecting \textit{unknown-unknown} to \textit{known-unknown}. How to leverage the availability of outliers is the key to success. Second, the classifier in CV and NLP is usually trained in a fully supervised manner with abundant labeled data, while for graphs the most common approach for node classification is to train a GNN with limited labeled data in a semi-supervised way. Due to the message-passing framework adopted by GNNs, the latent features of ID and OOD nodes can be affected by each other. Therefore, it is important to study how the information flow between inliers and outliers can affect the performance of in-distribution classification and outlier detection. A similar question arises in \cite{oliver2018realistic}, where the authors investigate the performance of semi-supervised learning (SSL) methods when labeled and unlabeled data are drawn from different distributions. However, this problem remains unexplored in the field of graph-based SSNC. Third, since our purpose is to address node classification and outlier detection in a joint framework, a natural question is how to combine the two tasks into a unified model, and how to balance the impact of one task on the other. 

In this work, we first analyze the impact of OOD nodes on graph learning tasks with GNNs. We demonstrate that for graphs with high homophily, message-passing GNNs are inherently good at detecting outliers due to the smoothness effect caused by feature propagation. Furthermore, we find that removing inter-edges between ID and OOD nodes while preserving intra-edges within each cluster can lead to the overall best performance. Motivated by these findings, we propose a novel GNN model Out-of-Distribution Graph Attention Network (OODGAT) which utilizes attention mechanism and explicitly models the interaction between inliers and outliers. Experiments show that OODGAT outperforms all baselines in terms of both detection and classification, and even surpasses post-hoc detectors which are tuned directly on the test set.

To the best of our knowledge, we are the first to formally define the problem of \textit{graph learning with OOD nodes}. \cite{uncertainty2020} considers a similar setting where the graph also contains OOD nodes. They developed a Bayesian framework to detect outliers by calculating multiple uncertainty measures. Our work differs in that we analyze the fundamental advantages of GNNs from the perspective of network geometry, and exploit the information contained in the graph structure to solve the problem in an efficient and elegant way.

To summarize, our work makes the following contributions:
\begin{itemize}
    \item We formalize the problem of \textit{graph learning with OOD nodes} and identify its challenges.
    \item We analyze the problem from the perspective of graph structure and present the basic design choice to achieve good performance.
    \item We propose a novel GNN model called OODGAT which explicitly distinguishes inliers from outliers during feature propagation and solves the problem of node classification and outlier detection in a joint framework.
    \item We conduct extensive experiments on various graph datasets to demonstrate the effectiveness of the proposed method. 
\end{itemize}

\section{Related Works}
\subsection{Graph Neural Networks}
Graph neural networks (GNNs) have shown great performance in various applications related to graphs. In this work, we focus on the problem of semi-supervised node classification (SSNC) \cite{kipf2016semi}. In SSNC, GNNs aggregate features from neighboring nodes and produce a latent space where the similarity between node embeddings corresponds to the connection patterns between nodes in the geometry space. The most commonly used GNNs include graph convolutional network (GCN) \cite{kipf2016semi}, graph attention network (GAT) \cite{gat} and GraphSAGE \cite{sage2017}. 

\subsection{Outlier Detection}
Outlier detection, also known as OOD detection,  has been a hot research area in various domains. Based on the availability of OOD data during training, OOD detectors can be classified into three types, namely unsupervised, supervised and semi-supervised methods. 

\noindent \textbf{Unsupervised Methods.} 
Unsupervised methods only utilize in-distribution data to train the outlier detector. Among various techniques, the most commonly used ones include ODIN \cite{odin2017} and Mahalanobis-distance \cite{m2018}. These methods are called \textit{post-hoc detectors} as they assume the classification network is already trained on in-distribution data, and the detector is built on top of the pretrained classifier by calibrating its output probabilities or exploiting its latent space. Other approaches like \cite{eloc2018, csi2020, ssd2021} require training an additional model which is designed specifically for OOD detection, apart from the original classification network. Unsupervised methods do not utilize the abundant unlabeled data during training and can only find sub-optimal solutions since they treat the classification and outlier detection as two independent tasks. 

 \noindent \textbf{Supervised Methods.} 
Supervised methods assume access to a set of OOD samples during training \cite{oe2018, acet2019, confidence2017}. Such methods train the classifier in an end-to-end fashion using cross-entropy loss on the ID training data to minimize classification error, together with a confidence penalty loss on the labeled OOD data to maintain low prediction confidence. For example, \cite{confidence2017} applies a KL-divergence term to OOD samples to ensure their predictions are close to uniform distribution. Supervised detectors generally outperform unsupervised ones for they manage to exploit the distributional information provided by training OOD data. However, the OOD samples are either from a different but related dataset \cite{oe2018} or generated by GANs \cite{oodgan2021}, limiting its application in the graph domain where one cannot find such proxy OOD datasets and cannot easily generate pseudo-OOD data.

 \noindent \textbf{Semi-supervised Methods.} 
Inspired by semi-supervised learning, recent studies in OOD detection also consider the setting where an unlabeled set is available during training \cite{step2021, mcd2019, tifrea2021novelty}. \cite{step2021} defines a novel task called 'semi-supervised OOD detection', where one is given a limited set of labeled inliers and a large mixed set of both inliers and outliers, whose identities cannot be known during training. They employ contrastive learning to obtain latent representations of unlabeled samples and compute their distance from the centers of in-distribution data as the OOD score. \cite{tifrea2021novelty} adopts a similar setting but solves the problem with ensemble. The drawbacks of these methods include 1) they are not designed for graphs and thus cannot utilize the structural information; 2) they usually require training an additional detection model and cannot handle classification and detection in the same framework.  

\subsection{Semi-supervised Learning With Distribution Mismatch}
Another way to understand the proposed task is to consider it as a semi-supervised learning problem on graphs. SSL assumes access to only a small set of labeled data and a relatively large set of samples without label information. Oliver et al. \cite{oliver2018realistic} points out that existing SSL methods tend to degrade the original classification performance when there exists a class distribution mismatch between labeled and unlabeled data. Following their discovery, researchers have developed SSL methods that are robust against OOD samples, with their performance being at least as good as fully-supervised learning \cite{guo2020safe, openssl2020, openmatch2021}. The key idea of such methods is straightforward: they attempt to detect and remove the OOD part of the unlabeled data and apply SSL techniques only on the remaining purified set. This setting resembles ours in that they also treat the problem as two tasks, i.e., semi-supervised learning on in-distribution data and outlier detection on the unlabeled set, where each task has its influence on the other. However, these approaches perform SSL by adding regularization terms to the original classification loss (e.g., cross-entropy), like VAT \cite{vat2018} and minimum entropy regularization \cite{entropy2005}, while in the graph domain, SSNC is usually done with GNNs which achieve semi-supervised learning in an implicit way.

\section{Learning on Graphs with Out-of-distribution Nodes}
\label{sec:analysis}
In this section, we define the problem of \textit{graph learning with OOD nodes} and discuss the design choice of OODGAT.

\subsection{Problem Formulation}
Consider a graph $\mathcal{G}=(\mathcal{V}, \mathcal{E})$, where $\mathcal{V}$ denotes the set of nodes and $\mathcal{E}$ denotes the set of edges. The graph structure is represented by an binary adjacency matrix $\mathbf{A}\in\{0,1\}^{|\mathcal{V}|\times|\mathcal{V}|}$. Each node $v$ in the graph is associated with a feature vector $\mathbf{x}_v$ and a label $y_v$, and the overall feature matrix and class vector can be represented by $\mathbf{X}$ and $\mathbf{y}$, respectively. In SSNC, the node set can be further divided into $\mathcal{V}=\mathcal{V}_l\cup \mathcal{V}_u$ where $\mathcal{V}_l$ refers to the set of nodes whose labels are accessible during training. Similarly, the feature matrix and class vector can be divided into 
$\mathbf{X}=
{\left[ 
    {\mathbf{X}_l}^\top,{\mathbf{X}_u}^\top
 \right]} ^\top$ 
and $\mathbf{y}=[\mathbf{y}_l\mathbin \Vert \mathbf{y}_u]$. The aim of SSNC is to predict the labels for nodes in $\mathcal{V}_u$ using the training set $(\mathbf{X}_l, \mathbf{y}_l)$, the unlabeled features $\mathbf{X}_u$ and the graph structure $\mathbf{A}$. Different from traditional close-world SSNC which assumes that nodes in $\mathcal{V}_l$ and $\mathcal{V}_u$ are sampled from the same distribution, we generalize the problem into a more realistic setting where nodes in $\mathcal{V}_u$ may come from a different distribution than nodes in $\mathcal{V}_l$. Due to the distribution shift between labeled and unlabeled data, the class vector $\mathbf{y}_u$ may contain labels not seen in $\mathbf{y}_l$, and the label space $\mathcal{Y}$ is enlarged by $\mathcal{Y}=\mathcal{Y}_l\cup \mathcal{Y}_u$. For simplicity, we denote nodes with labels from $\mathcal{Y}_l$ by ID nodes or inliers and nodes with labels from $\mathcal{Y}_u\setminus\mathcal{Y}_l$ by OOD nodes or outliers. We call this setting \textit{graph learning with OOD nodes} and the purpose is to 1) assign labels from $\mathcal{B} = \{0, 1\}$ to nodes in $\mathcal{V}_u$ where 0 stands for ID and 1 for OOD and 2) for nodes tagged as ID, we further classify them to be one of the classes in $\mathcal{Y}_l$. Note that for both tasks, we are presented with the whole graph $\mathcal{G}$ during training, leading to a \textit{semi-supervised} and \textit{transductive} setting. In the remaining of the article, we call the two tasks Semi-Supervised Outlier Detection (SSOD) and Semi-Supervised Node Classification (SSNC) for the sake of simplicity. 

\subsection{Semi-supervised Outlier Detection}
Unlike previous outlier detection methods which are designed primarily for CV and NLP tasks and derive the detector using only in-distribution data, the uniqueness of graphs makes us wonder: \textit{can we leverage the unlabeled data $\mathbf{X}_u$ and the graph structure $\mathbf{A}$ for better OOD detection?} 
To answer the question, we first take a brief review at the most common task on graphs, namely, SSNC. In SSNC, a GNN is used to propagate information between adjacent nodes and produce a latent space where features are distributed smoothly w.r.t. the graph structure \cite{insight2018}. The smoothness is desirable due to the widely adopted homophily assumption, i.e., connected nodes tend to share the same label \cite{pei2020geom}. We argue that, like SSNC, the connection pattern between nodes can also provide information for distinguishing ID from OOD. We start by giving the following proposition:

\begin{prop}
Given a graph $\mathcal{G}$, the set of original labels $\mathcal{Y}=\mathcal{Y}_l\cup \mathcal{Y}_u$, and the set of identity labels $\mathcal{B}=\{0, 1\}$. Assume that:

(1) There exists a mapping $f: \mathcal{Y}\mapsto\mathcal{B}$ which maps each label in $\mathcal{Y}$ to be ID or OOD;

(2) $\mathcal{G}$ is homophilic w.r.t. to $\mathcal{Y}$, i.e., edges in $\mathcal{G}$ tend to connect nodes with the same label in $\mathcal{Y}$.

Then, $\mathcal{G}$ is also homophilic w.r.t. $\mathcal{B}$.
\end{prop}

\begin{figure}
\centering
    \begin{subfigure}{\linewidth}
        \centerline{\includegraphics[width=\linewidth, scale=0.5]{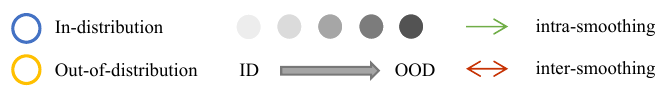}}
    \end{subfigure}
    
    \begin{subfigure}[b]{0.35\linewidth}  
        \centerline{\includegraphics[scale=0.50]{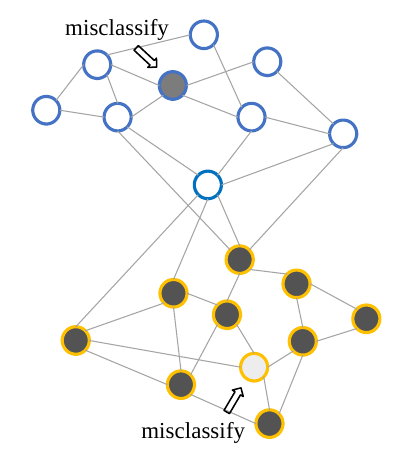}}
        \caption{Before smoothing}
    \end{subfigure}
    \hfill
    \begin{subfigure}[b]{0.35\linewidth}  
        \centerline{\includegraphics[scale=0.50]{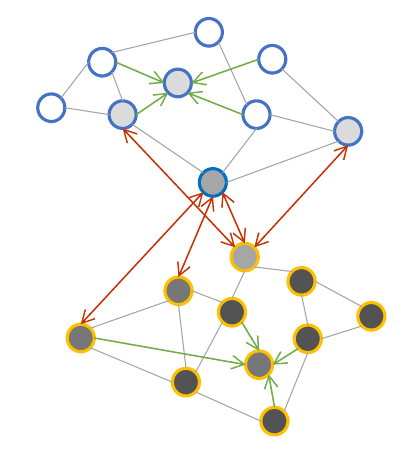}}
        \caption{After smoothing}
        \label{fig: smooth_b}
    \end{subfigure}
    \setlength{\belowcaptionskip}{-12pt}
    \caption{Smoothness helps OOD detection. The border of circles represents the true identity of nodes, while the darkness of the inner color represents the predicted OOD score. Arrows indicate the smoothing effect of GNNs.}
    \label{fig:smooth}
    \Description{}
\end{figure}

The proof of the proposition is presented in Appendix \ref{sec:proof}. 
From the proposition, we make the following hypothesis: 
GNNs are a natural fit for SSOD because they are inherently equipped with a regularizer that pushes the predicted OOD scores to be close within densely connected communities, which is helpful for graphs with high homophily. We illustrate this in Figure \ref{fig:smooth}. The left figure shows the OOD scores obtained without considering the graph structure. Overall, the scores of OOD nodes are higher than ID nodes, with an exception in each community due to the weakness of modern neural networks \cite{acet2019}. By smoothing features according to the graph structure (Figure \ref{fig: smooth_b}), GNN manages to recover the true scores of nodes from their neighbors (green arrows). However, we also notice the edges that connect different kinds of nodes, which lead to undesirable feature aggregation and compromise the separation between inliers and outliers (red arrows). Since the number of intra-edges significantly exceeds that of inter-edges (for graphs with high homophily), the overall performance should be better than not utilizing structural information at all.

\begin{figure}
\centering
    \begin{subfigure}[b]{0.40\linewidth}  
        \centerline{\includegraphics[scale=0.35]{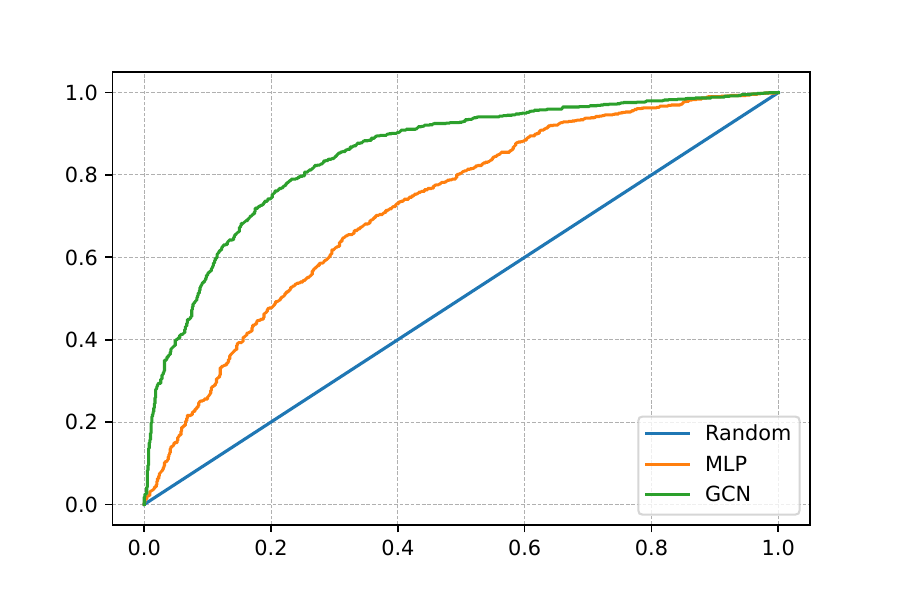}}
        \caption{ROC curves of OOD detection using MLP and GCN as detectors. GCN uses the original graph as input.}
        \label{fig:roc_a}
    \end{subfigure}
    \hfill
    \begin{subfigure}[b]{0.45\linewidth}  
        \centerline{\includegraphics[scale=0.35]{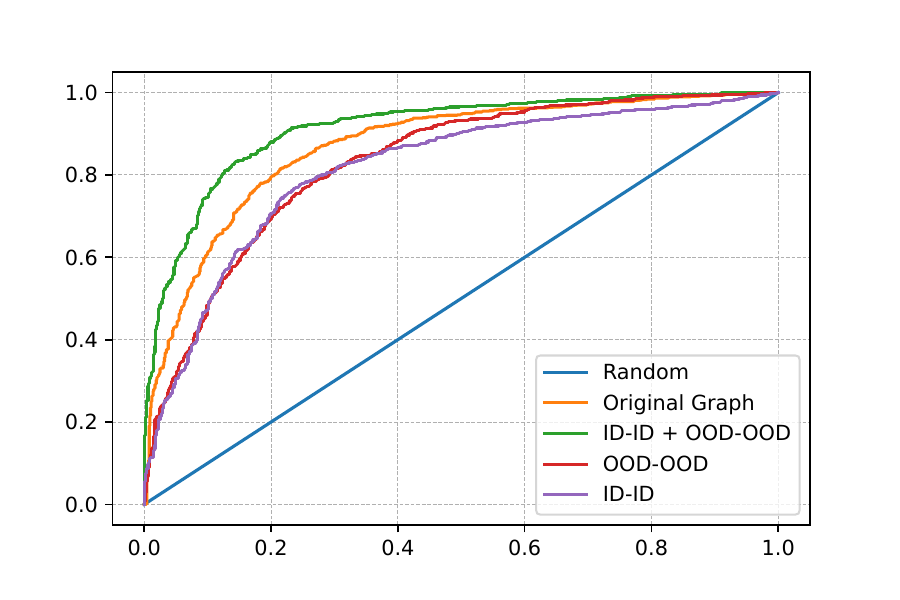}}
        \caption{ROC curves of GCN detectors. Different lines are trained on graphs with different subsets of edges.}
        \label{fig:roc_b}
    \end{subfigure}
    \caption{Effect of feature smoothing on Cora. }
    \Description{}
\end{figure}

To verify the hypothesis, we conduct an experiment on Cora \cite{cora2016} using Multilayer Perceptron (MLP) and GCN \cite{kipf2016semi} as predictors, and calculate the entropy of the predicted class distribution as the OOD score as in \cite{likelihood2019, uncertainty2020, entropic2021}. The higher the entropy, the more likely the model considers the node to be OOD. The ROC curves for both methods are plotted in Figure \ref{fig:roc_a}, from which we can see the GCN detector outperforms the MLP counterpart by a large margin, validating that the graph structure is useful for detecting outliers. To better understand the impact of different kinds of connections, we test the detection performance on graphs with different subsets of edges, and the results are shown in Figure \ref{fig:roc_b}. As expected, removing inter-edges from the graph leads to improved detection performance (green line vs. orange line). However, the performance drops sharply when we further remove edges within ID (red line) or OOD (purple line) communities, indicating that smoothness within the same type of nodes is critical for successful detection.
\subsection{Semi-supervised Node Classification}
It is known that the distribution mismatch between labeled and unlabeled data can hurt the performance of semi-supervised learning \cite{oliver2018realistic}. In graph-based SSNC, unlabeled nodes convey their influence to model parameters through their connections to the labeled ones, so it is natural to expect the same performance drop observed in \cite{oliver2018realistic} when the graph contains edges connecting inliers and outliers. However, the problem here is more sophisticated. On the one hand, the information exchange between ID and OOD data may introduce noise to the interested distribution, making the model prone to overfitting and leading to poor generalization; On the other hand, the addition of inter-connections can enhance the connectivity of the graph and facilitate the propagation of supervision signals among nodes. Moreover, the connection patterns between inliers and outliers may provide knowledge about how to classify ID nodes. Therefore, it is difficult to tell whether the presence of inter-connections is beneficial or detrimental to SSNC. To find out the impact of inter-connections, we conducted experiments on
some commonly used graph datasets using GCN as the classifier and report the mean accuracy across 9 runs in Table \ref{tab:edge_removal}. For each graph, we test the classification accuracy in three cases: preserving all inter-edges (remove=0), randomly dropping half of them (remove=0.5), and removing them all (remove=1.0). Empirically, removing inter-edges can improve the generalization of in-distribution classification, which is particularly true for certain datasets.

\begin{table}
  \caption{Effect of Inter-Edge Removal}
  \label{tab:edge_removal}
  \begin{tabular}{c|cccc}
    \toprule
    Dataset & Remove=0 & Remove=0.5 & Remove=1.0\\
    \midrule
    Cora & 92.0 & 92.5 & \bftab 92.7\\
    CoauthorCS & 92.8 & 92.6 & \bftab 93.0\\
    Amazon-Photo & 97.0 & 97.0 & \bftab 97.2\\
    Amazon-Computers & 81.2 & 81.5 & \bftab 83.2\\
    \bottomrule
  \end{tabular}
\end{table}

\section{OODGAT: End-to-end Model for SSOD and SSNC}
In this section, we first introduce the attention architecture adopted by OODGAT, and then propose three regularization terms to guide the learning process of OODGAT. 
\subsection{Attention Mechanism: From Node to Edge}
Since it is important to separate in-distribution nodes from OOD nodes, it is natural to resort to attention mechanism which adaptively computes the weights for aggregating information from neighbors. The general form of graph convolution with attention is:
\begin{equation}
    \mathbf{h}_i^{\prime}=\sigma\left(\sum_{j\in \mathcal{N}(i) \cup \{v_i\}} {\alpha_{ij} \mathbf{W} \mathbf{h}_j} \right)
\end{equation}
where $\alpha_{ij}$ is the attention weight for aggregating information from $v_j$ to $v_i$. The difference between various graph attention networks lies in the way the attention values are calculated. For example, GAT \cite{gat} proposes to compute the (unnormalized) attention weights between $v_i$ and $v_j$ by $e_{ij}=LeakyReLU\left(\mathbf{a}^\top \left[\mathbf{W}\mathbf{h}_i \mathbin \Vert \mathbf{W}\mathbf{h}_j\right]\right)$, where they use a single layer neural network parameterized by $\mathbf{a}$ to output attention weights. However, none of the previous approaches takes OOD nodes into account, and the attention coefficients obtained from their methods are not guaranteed to contain knowledge about how to distinguish inliers from outliers.  

\begin{figure}
     \centering
     \includegraphics[scale=0.40]{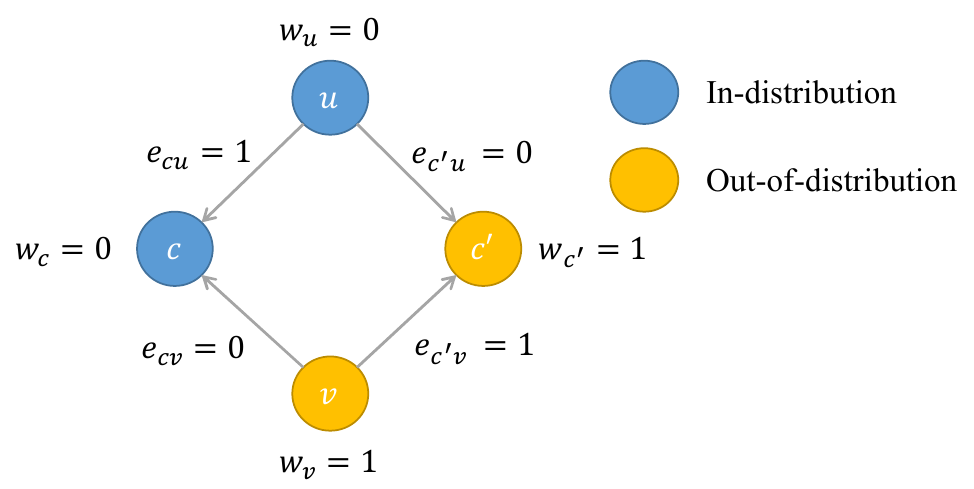}
     \caption{Attention computation of OODGAT. $c$ and $c^\prime$ indicate central nodes, while $u$ and $v$ are their neighbors. Arrows indicates the direction of feature propagation. OODGAT first computes the node-level OOD scores, and then converts the scores into edge-level attention weights for feature propagation. }
     \label{fig:oodgat}
 \end{figure}

In OODGAT, we propose to explicitly model the interaction between inliers and outliers. Based on the discussion in Section \ref{sec:analysis}, we summarize three properties the attention mechanism should possess : 1) allow messages to pass within in-distribution nodes, 2) allow message passing within out-of-distribution nodes and 3) block information flow between inliers and outliers. Therefore, we propose the following attention form:
\begin{equation}
    e_{ij}=1-\left|w(i)-w(j)\right|
    \label{equ:unormalized}
\end{equation}
 where $w(i)$ and $w(j)$ are the attention scores for $v_i$ and $v_j$, respectively. If we consider $w(v)$ as a binary classifier that assigns different weights to inliers and outliers, we can find that Equation (\ref{equ:unormalized}) satisfies all the properties discussed above. We illustrate this in Figure \ref{fig:oodgat}. Without loss of generality, when $w_v$ and $w_{c^\prime}$ are large, say $w_v=w_{c^\prime}=1$, and $w_u$ and $w_c$ are small, say $w_u=w_c=0$, the attention weights for intra-edges become $e_{cu}=e_{c^\prime v}=1$, while the weights for inter-edges become $e_{cv}=e_{c^\prime u}=0$. We also note that, for any node $v_i$, the weight with which it attends to itself is fixed to be $e_{ii}=1$, i.e., the maximum value possible for all node pairs. This is desirable since maintaining more self-information can be helpful when the neighborhood may contain contaminated features. After obtaining $e_{ij}$, we normalize them in each neighborhood using softmax to keep the embedding scale unchanged before and after aggregation:
 \begin{equation}
     \alpha_{ij} = softmax_j (e_{ij}) = \frac{exp(e_{ij})}{\sum_{k\in\mathcal{N}(i)\cup\{v_i\}} exp(e_{ik})}
 \end{equation}
 
 The binary classifier $w(v)$ can be defined in various forms. To avoid too many parameters and a complex model, we simply implement it as a logistic regression classifier parameterized by $\mathbf{a}\in \mathbb{R} ^{d^\prime}$ over the latent space of a GNN layer, i.e., $w(v)=\sigma\left(\mathbf{a}^\top\mathbf{W}\mathbf{h}_v\right)$, where $\mathbf{W}\in\mathbb{R}^{d^{\prime}\times d}$ is the weight matrix of the GNN layer, $\mathbf{h}_v\in\mathbb{R}^d$ is the input of the layer, and $\sigma$ is the sigmoid function. The classifier aims to find a partition of the latent space such that inliers and outliers are well separated from each other. To enhance the expressiveness of the model, we extend the attention computation to a multi-head variant, similar to \cite{gat}:
 \begin{equation}
     \mathbf{h}_i^{\prime} = \bigparallel_{k=1}^K \sigma \left(\sum_{j \in \mathcal{N}(i) \cup \{v_i\}} \alpha_{ij}^k \mathbf{W}^k \mathbf{h}_j\right)
 \end{equation}
 where $K$ is the number of attention heads and $\bigparallel$ means concatenation. In the prediction layer, the concatenation is replaced with averaging to keep the dimension reasonable for classification:
  \begin{equation}
     \mathbf{z}_i =  softmax \left(\frac{1}{K} \sum_{k=1}^K \sum_{j \in \mathcal{N}(i) \cup \{v_i\}} \alpha_{ij}^k \mathbf{W}^k \mathbf{h}_j\right)
 \end{equation}

 where $\mathbf{z}_i$ is the predicted class distribution of $v_i$, $\sum_{k=1}^{|\mathcal{Y}_l|} z_{ik}=1$. 
 
\subsection{Regularizer}
\label{sec:reg}
\begin{figure}
    \centering
    \includegraphics[scale=0.37]{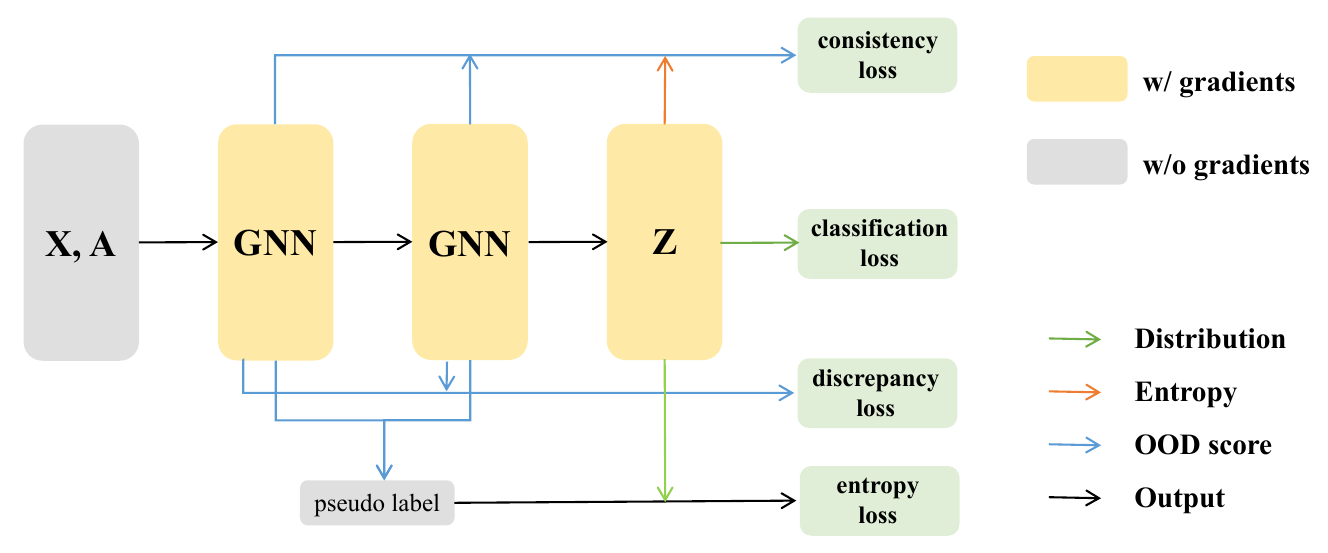}
    \caption{Architecture of OODGAT. 
    Arrows of different colors indicate different information to be extracted from the layer for loss computation. Yellow and grey rectangles represent layers w/ and w/o gradients propagation, respectively.}
    \label{fig:architecture}
\end{figure}

Of course, the cross-entropy loss alone is not sufficient to make OODGAT work in the expected way. In particular, we want the embedded binary classifier to learn knowledge about how to distinguish inliers from outliers. Therefore, we propose three regularizers to guide the learning process of OODGAT, i.e., consistency loss, entropy loss and discrepancy loss. The architecture of OODGAT is shown in Figure \ref{fig:architecture}, which we will explain in detail in the following sections.

\noindent \textbf{Consistency Regularizer.} 
OODGAT integrates a binary classifier to measure the OOD score for nodes in the graph, and translates the node-level scores into edge-level attention weights for feature aggregation. Besides the scores predicted by the classifier, we can also obtain the output distribution of nodes at the final layer of the model, from which the entropy can be calculated as another kind of OOD measurement. We denote the scores predicted by the classifier as $w$, and the scores given by entropy as $e$. To coordinate the relationship between $w$ and $e$, we design the following regularization term called consistency loss:
\begin{equation}
    \mathcal{L}_{con} =-\cos\left(\mathbf{w},\mathbf{e}\right)
    \label{eq:consistency}
\end{equation}
where $\mathbf{w}$ represents the vector of OOD scores predicted by the classifier:
\begin{equation}
\begin{aligned}
    \mathbf{w} = [w_1, w_2, & \cdots, w_{|\mathcal{V}|}]^\top\\
    w_i = \sigma &(\mathbf{a}^\top \mathbf{Wh}_i)\\
\end{aligned}
\end{equation}
and $\mathbf{e}$ denotes the vector of OOD scores given by entropy:
\begin{equation}
\begin{aligned}
    \mathbf{e} = [\sigma(\tilde{e_1}), \sigma(&\tilde{e_2}), \cdots, \sigma(\tilde{e_{|\mathcal{V}|})}]^\top\\
    \tilde{e_i} = &\frac{e_i-\mu_e}{\sigma_e}\\
    e_i = H(\mathbf{z}_i) =& - \sum_{j=1}^{|\mathcal{Y}_l|} z_{ij}log(z_{ij})
\end{aligned}
\end{equation}
where $\mu_e$ and $\sigma_e$ denotes the mean and standard deviation of ${\{e_i\}}_{i=1}^{|\mathcal{V}|}$, and $H(\mathbf{z})$ is the entropy of the predicted class distribution given by the last layer of OODGAT. 

In Equation (\ref{eq:consistency}), we use cosine similarity to constrain the difference between $\mathbf{w}$ and $\mathbf{e}$, i.e., the two methods should give similar predictions across all nodes.  
The intuition behind the consistency regularizer is the causal relationship between the attention mechanism and the model's final output. That is, when the scores given by the classifier change, the attention weights used for aggregating features will also change, which in turn affects the final output of the model. If we regard the change of the classifier as the cause, then the change of the model's output can be viewed as the effect. By aligning cause and effect, the hypothesis space of the model is reduced and gradient descent is more likely to find solutions that are closer to ground truth. Imagine an extreme case where the classifier works perfectly and produces OOD scores close to ground-truth. In this case, the attention weights of edges also become near perfect and the model becomes extremely powerful in detecting outliers as it smooths representations for all ID and OOD clusters and prevents the information exchange between ID and OOD communities. As a result, the OOD scores computed from entropy are also close to reality, making the angle between $\mathbf{w}$ and $\mathbf{e}$ small.
From another perspective, we can interpret the consistency loss as a kind of supervised learning: the entropy provides supervision to the classifier and vice versa. As training progresses, the classifier not only learns from the final output, but also teaches the model to produce more reliable predictions by differentiating ID and OOD better in the latent space. Thus, the two modules play a chasing game and benefit each other.

For OODGAT with two layers, the consistency loss is computed for both layers, and in each layer, the score vector $\mathbf{w}$ is averaged across all heads:
\begin{equation}
\begin{aligned}
    \mathcal{L}_{con} & = -\frac{1}{2} \times \left(\cos(\mathbf{w}^1,\mathbf{e}) + \cos(\mathbf{w}^2,\mathbf{e})\right)\\
    \mathbf{w^l} = \biggl[\frac{1}{K}&\sum_{k=1}^{K} w_1^{lk}, \frac{1}{K}\sum_{k=1}^{K} w_2^{lk}, \cdots, \frac{1}{K}\sum_{k=1}^{K} w_{|\mathcal{V}|}^{lk} \biggr]^\top
\end{aligned}
\end{equation}

\noindent \textbf{Entropy Regularizer.}  
In OODGAT, we use entropy as the measure of predictive uncertainty. As training proceeds, the cross-entropy loss continuously pulls the outputs of labeled nodes toward one-hot distribution, pushing their entropy to the lowest level. Due to the generalization ability of neural networks, nodes outside the training set may also produce low-entropy predictions, especially those with attributes similar to or closely connected to the labeled ones, resulting in some low-entropy regions in the graph. In contrast to the classification loss, we want to keep the uncertainty of outliers as high as possible to counteract the entropy-reducing effect caused by cross-entropy. However, the true identities of nodes cannot be obtained during training, so we take the predictions given by the binary classifier as pseudo-labels, and make the outputs of pseudo-OOD nodes close to uniform distribution to enhance the distinguishability between inliers and outliers. Thus, we get the following entropy loss:
 \begin{equation}
     \mathcal{L}_{ent}=\frac{\sum_{i=1}^{\left|\mathcal{V}\right|}CE\left(\mathbf{u},\mathbf{z}_i\right)\delta\bigl(w(i)>\epsilon\bigr)}{\sum_{i=1}^{\left|\mathcal{V}\right|}\delta\bigl(w(i)>\epsilon\bigr)}
 \end{equation}
 where $\mathbf{u}$ is uniform distribution, $\mathbf{z}_i$ is the predicted class distribution of $v_i$, $\epsilon$ is the threshold for selecting pseudo-OOD nodes, $\delta$ means the Kronecker delta. 
 
\begin{figure}
\centering
    \begin{subfigure}{\linewidth}
        \centering
        \includegraphics[scale=0.35]{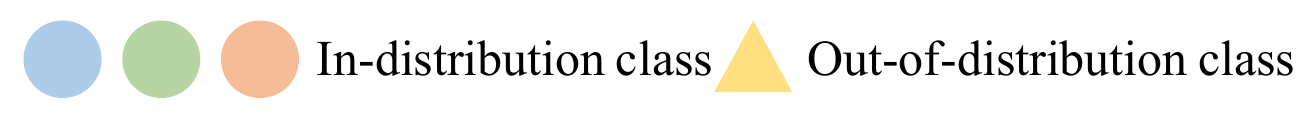}
    \end{subfigure}
    
    \begin{subfigure}[b]{0.45\linewidth}  
        \centering
        \includegraphics[scale=0.2]{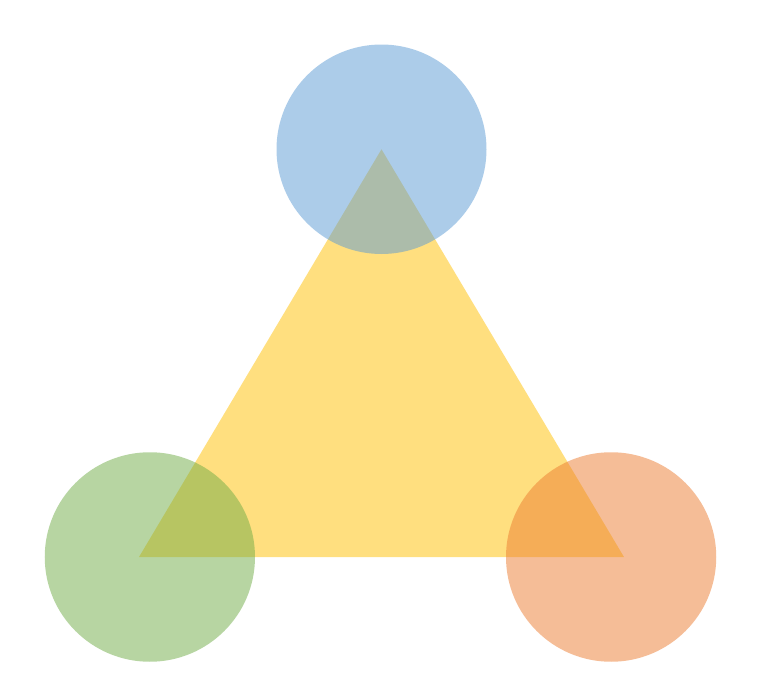}
        \caption{Ordinary GNNs}
    \end{subfigure}
    \hfill
    \begin{subfigure}[b]{0.45\linewidth}  
        \centering
        \includegraphics[scale=0.2]{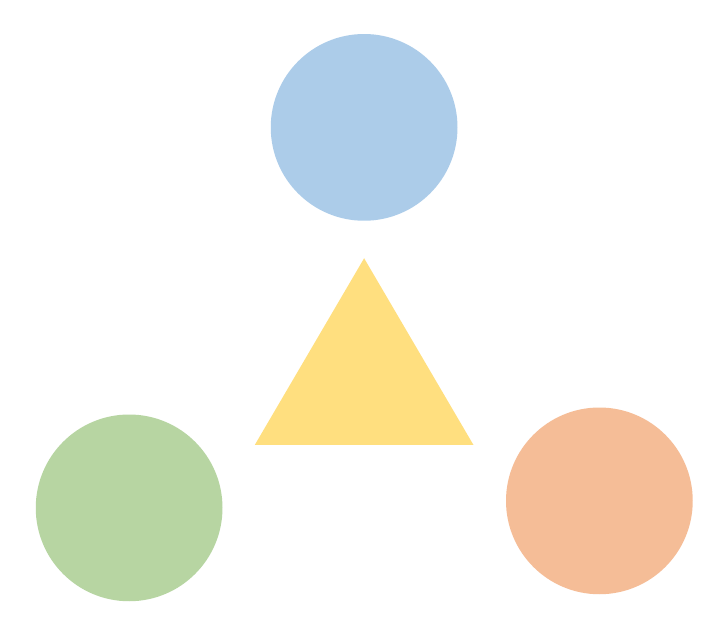}
        \caption{OODGAT}
    \end{subfigure}
    \caption{Illustration of the latent space of ordinary GNNs and OODGAT. The proposed regularizers can help the attention module to control the information flow between ID and OOD nodes, hence OODGAT is able to produce a clearer gap between inliers and outliers.}
    \label{fig:space}
    \Description{Entropy propagates from training nodes to the rest of the graph.}
\end{figure}

\noindent \textbf{Discrepancy Regularizer.} 
For OODGAT with two graph convolutional layers, we further constrain the difference between the OOD scores computed by the two layers by minimizing the following discrepancy loss:
 \begin{equation}
     \mathcal{L}_{dis}=-\cos\left(\mathbf{w}^1,\mathbf{w}^2\right)
 \end{equation}
\noindent \textbf{Final Objective.}
Overall, the optimal parameters of OODGAT are obtained by minimizing the following loss:
\begin{equation}
\begin{aligned}
        \mathcal{L}_{OODGAT} &=- \frac{1}{|\mathcal{V}_l|} \sum_{i=1}^{|\mathcal{V}_l|} log(z_{iy_i})  \\
        & +a^{b\times t}(\beta \mathcal{L}_{con}
        + \gamma \mathcal{L}_{ent}
        + \zeta \mathcal{L}_{dis})
\end{aligned}
\end{equation}
where $\beta$, $\gamma$ and $\zeta$ are the balance parameters of regularizers. In addition, $a^{b\times t}$ is used to decay the weights of regularizers gradually as training progresses so as to control the balance point between ID classification and OOD detection. $a$ is a number between 0 and 1, $b$ is a small number and $t$ is the iteration step. In the experiments, we set $a$ and $b$ to be 0.9 and 0.01, respectively. 
By combining the three regularizers with cross-entropy, OODGAT not only learns to classify in-distribution nodes, but also to separate inliers from outliers in the latent space, as shown in Figure \ref{fig:space}. 

\section{Experiments}
In this section, we test OODGAT on various real-world graph datasets to demonstrate its effectiveness. Due to space constraints, some visualizations are presented in Appendix \ref{sec:visual}. 
\subsection{Experimental Setup}
\noindent \textbf{Evaluation Metrics.}
In the setting of \textit{graph learning with OOD nodes}, we aim to accomplish two tasks simultaneously, which are 1) node classification and 2) outlier detection. For the first task, we adopt classification accuracy as the evaluation metric. For the second task, we calculate two metrics commonly found in the OOD detection literature, namely the area under ROC curve (AUROC) and the false positive rate when the true positive rate achieves 95\% (FPR@95). Note that in all experiments we view the outliers as positive. To comprehensively evaluate the performance of the two tasks, we consider them together as a multi-class classification problem with N+1 classes, i.e., N in-distribution classes and one OOD class. We call this task \textit{joint classification}, and the performance can be evaluated by weighted-F1\footnote{The details of joint classification are explained in Appendix \ref{sec:joint}.}.

\noindent \textbf{Datasets.}
We test OODGAT on six commonly used graph datasets, i.e., Cora, AmazonComputers, AmazonPhoto, CoauthorCS, LastFMAsia and Wiki-CS \cite{cora2016, amazon2018, lastfm2020, wiki2020}. For each dataset, we divide all classes into in-distribution and out-of-distribution, such that the ID part contains classes that are relatively balanced in node size, and the number of ID classes is at least three to avoid too easy classification\footnote{See our Github for more details regarding the choice of ID and OOD classes.}. Similar to traditional SSNC, we randomly select 20 nodes per ID class as the training set. Besides, we construct a small validation set which contains 10 nodes from each ID class, and the same number of outliers randomly sampled from OOD classes. Statistics for the datasets are listed in Appendix \ref{sec:dataset}.

\noindent \textbf{Methods.}
We compare the following methods:
\begin{itemize}
    \item \textbf{End-to-end Methods}, which accomplish SSOD and SSNC in the same framework. Specifically, we choose MLP, GCN \cite{kipf2016semi}, GraphSAGE \cite{sage2017}, GAT \cite{gat}, and GATv2 \cite{gatv2} as the end-to-end baselines. MLP is used to test the performance without considering graph topology, while the other four are representative GNN models w/ or w/o graph attention. For all methods, we use the entropy of the predicted distribution as the OOD score.
    \item \textbf{Post-hoc OOD Detectors}, which require training an additional outlier detector on top of the pretrained classifier. We employ ODIN \cite{odin2017}, Mahalanobis-distance \cite{m2018}, and CaGCN \cite{cagcn2021} as the post-hoc detectors. ODIN uses temperature scaling and input preprocessing to calibrate the output distribution, while Mahalanobis-distance leverages the latent space of the pretrained classifier to compute the distance between testing samples and known inliers.  For each method, we use the metric described in the original paper for OOD detection, i.e., MSP for \cite{odin2017} and Mahalanobis-distance for \cite{m2018}. CaGCN is a recently published method for calibrating the output confidence of GNNs. Intuitively, we can use the calibrated confidence as the score for outlier detection.
    \item \textbf{GKDE} \cite{uncertainty2020},  the abbreviation for Graph-based Kernel Dirichlet distribution Estimation, a method specifically designed to detect outliers on graphs. It proposes a multi-source uncertainty framework using various types of predictive uncertainties from both deep learning and belief theory, and
    shows that \textit{vacuity} is the best metric for OOD detection.
    \item \textbf{OODGAT}, the method proposed in this paper. It has two versions: OODGAT-ENT which uses the entropy of the predicted distribution as the measure of outliers and OODGAT-ATT which uses the score given by the binary classifier instead. 
\end{itemize}

\noindent \textbf{Implementation Details.} 
For all graphs, we perform 3 random splits to obtain training, validation, and test sets. For each split, we initialize the model with 3 random seeds.\footnote{In the comparison with GKDE, we use the split given by the original authors in
\url{https://github.com/zxj32/uncertainty-GNN}} As a result, each experiment was performed 9 times.
Unless specially mentioned, we tune the hyperparameters using grid search and select the best performing results according to the validation set. Specifically, we choose the learning rate from [0.01, 0.1], the dropout probability from [0, 0.5]. For models with multi-head attentions, the number of attention heads is chosen from [1, 4, 8], and the drop edge probability is set to 0.6. It is known that weight decay is helpful in preventing models from giving arbitrary high confidence, so we also choose the weight decay factor from [0, 5e-5, 5e-4, 5e-3]. We set the maximum iterations of training to be 1000 and perform early-stopping when $(AUROC+Accuracy)$ stops to increase for 200 epochs. All experiments are done using PyTorch Geometric, and the source code is made publicly available on Github\footnote{\url{https://github.com/SongYYYY/KDD22-OODGAT}}.

\subsection{Main Results}
\noindent \textbf{Comparison with End-to-end Methods.}
We first compare our method with end-to-end approaches.
The results are listed in Table \ref{tab:performance}. From the table, we make the following observations:

1) On all datasets, GNNs outperform MLP in both SSOD and SSNC by a large margin, suggesting that the graph structure is helpful for both tasks, as indicated in Section \ref{sec:analysis}.

2) GraphSAGE surpasses GCN in terms of AUROC on 5 out of the 6 datasets, which may be attributed to the strategy of separating self and neighboring representations during feature propagation. 

3) Across all baseline models and datasets, GAT and/or GATv2 achieve the best performance in outlier detection. The results show that even the naive attention mechanism helps to distinguish nodes from different distributions.

4) For SSOD, OODGAT outperforms all baselines on the six datasets by a considerable margin. On easy datasets such as AmazonPhoto and CoauthorCS, OODGAT achieves an AUROC of over 0.98, while for difficult tasks like LastFMAsia and Wiki-CS, OODGAT greatly improves the detection ability and achieves decent performance, demonstrating the effectiveness of the proposed propagation strategy. 

5) For SSNC, OODGAT achieves better or comparable results than other approaches. For example, OODGAT outperforms GAT and/or GATv2 by 3\% and 1\% in terms of classification accuracy on AmazonComputers and LastFMAsia. 
 The results show that by removing the interference brought by OOD data, the classifier is more likely to converge to points with better generalization ability.
 
6) From the perspective of \textit{joint classification}, OODGAT consistently outperforms all competitors, making it the most powerful method for \textit{graph learning with OOD nodes}. 

\begin{table*}
  \caption{Comparison with End-to-end Methods}
  \label{tab:performance}
  \resizebox{\textwidth}{!}{%
  \begin{tabular}{ccccccc}
    \toprule
    & Cora & AmazonCS & AmazonPhoto & CoauthorCS & LastFMAsia & Wiki-CS\\
    \midrule
    & \multicolumn{6}{c}{\textcolor{blue}{Acc} $\uparrow$ \hspace{1mm}/ \textcolor{orange}{AUROC} $\uparrow$\hspace{1mm}/ \textcolor{olive}{FPR@95} $\downarrow$\hspace{1mm}/ \textcolor{teal}{F1} $\uparrow$}\\
    \hline
    MLP & 74.1/72.4/75.5/63.1 &
    68.4/65.7/84.6/54.6 &
    91.8/80.2/71.9/72.8 &
    88.6/95.0/28.9/84.8 &
    54.5/57.4/87.0/51.2 &
    78.6/71.7/76.4/64.0\\
    GCN \cite{kipf2016semi} & 92.1/88.9/46.0/80.5 &
    81.2/83.3/61.9/70.3 &
    97.1/88.3/44.6/80.7 &
    \textcolor{blue}{\textbf{92.7}}/94.5/32.2/86.4 &
    79.8/72.1/74.7/66.5 & 
    80.9/71.7/76.6/63.0\\
    SAGE \cite{sage2017} & 90.8/87.7/46.6/79.2 &
    83.2/84.6/54.9/71.7 &
    97.1/93.5/32.0/87.2 &
    92.6/97.0/16.8/89.1 &
    79.3/73.7/68.9/67.0 &
    78.6/73.0/65.3/66.2\\
    GAT \cite{gat} & 91.6/90.1/40.8/81.5 &
    82.3/88.5/42.9/76.5 &
    96.9/92.5/31.7/86.1 &
    92.0/96.6/16.7/89.0 &
    82.3/81.1/49.6/75.0 &
    79.9/79.8/63.6/70.0\\
    GATv2 \cite{gatv2} & 91.5/90.4/40.0/81.9 &
    83.5/88.6/45.7/76.3 & 
    95.7/94.4/21.1/88.4 &
    91.7/96.6/19.1/88.7 &
    81.9/79.7/52.3/73.5 &
    \textcolor{blue}{\textbf{81.4}}/80.9/58.9/70.6\\
    \hline
    OODGAT-ENT & 
    \textcolor{blue}{\textbf{92.3}}/92.9/31.4/84.4 & 
    \textcolor{blue}{\textbf{86.6}}/92.2/\textcolor{olive}{\textbf{40.4}}/81.4 & 
    \textcolor{blue}{\textbf{97.6}}/98.2/8.1/93.4 & 
    92.4/98.9/3.7/92.6 &  
    \textcolor{blue}{\textbf{83.3}}/\textcolor{orange}{\textbf{93.4}} /\textcolor{olive}{\textbf{22.4}}/\textcolor{teal}{\textbf{83.5}} & 
    \textcolor{blue}{\textbf{81.4}}/\textcolor{orange}{\textbf{88.8}}/\textcolor{olive}{\textbf{48.5}} /\textcolor{teal}{\textbf{76.6}}\\
    OODGAT-ATT & 
    \textcolor{blue}{\textbf{92.3}}/\textcolor{orange}{\textbf{93.6}}/\textcolor{olive}{\textbf{26.1}} /\textcolor{teal}{\textbf{85.1}} &
    \textcolor{blue}{\textbf{86.6}}/\textcolor{orange}{\textbf{93.1}}/45.2/\textcolor{teal}{\textbf{82.2}} &
    \textcolor{blue}{\textbf{97.6}}/\textcolor{orange}{\textbf{98.3}}/\textcolor{olive}{\textbf{5.8}} /\textcolor{teal}{\textbf{93.9}} &
    92.4/\textcolor{orange}{\textbf{99.6}}/\textcolor{olive}{\textbf{1.6}} /\textcolor{teal}{\textbf{93.5}} &
    \textcolor{blue}{\textbf{83.3}}/91.9/27.7/81.0 & 
    \textcolor{blue}{\textbf{81.4}}/88.3/51.2/73.7\\
    \bottomrule
  \end{tabular}}
\end{table*}

\noindent \textbf{Comparison with Post-hoc OOD Detectors.}
We also compare OODGAT with ODIN \cite{odin2017}, Mahalanobis-distance \cite{m2018} and CaGCN \cite{cagcn2021}. The comparison is unfair as these methods either require additional data preprocessing or involve multiple training stages, while OODGAT accomplishes the mission without introducing additional complexity.  
For all experiments except OODGAT, we pretrain a GAT as the base classifier, and employ different post-hoc detectors for OOD detection. Note that unlike the original paper, we tune the detectors directly on the test set to eliminate the possibility of bad hyperparameter configurations. For OODGAT, we do not utilize the test set for training or tuning. Table \ref{tab:performance_2} reports the detection performance of all methods. 
As we can see, only in a few cases can the post-hoc detectors improve the detection ability (shaded cells). Apart from that, all methods lose their power due to the characteristics of graph data such as lack of supervision and non-continuous input.
By inspecting the last two columns, we find that despite being unfair, OODGAT outperforms all post-hoc detectors by a large margin. The superiority of OODGAT comes from the end-to-end optimization strategy  which simultaneously handles feature extraction and OOD detection, whereas other methods use a two-stage update framework  which trains the classifier and the detector separately and can only find sub-optimal solutions. 

\begin{table*}
\begin{minipage}{0.60\textwidth}
  \centering
  \caption{Comparison with Post-hoc OOD Detectors}
  \label{tab:performance_2}
  \resizebox{\linewidth}{!}{%
  \begin{tabular}{ccccccc}
    \toprule
    & GAT & ODIN & Mahalanobis & CaGCN & OODGAT & OODGAT\\
    & (base)\cite{gat} & \cite{odin2017} & -Distance\cite{m2018} & \cite{cagcn2021} & -ENT & -ATT\\
    \midrule
    & \multicolumn{6}{c}{\textcolor{orange}{AUROC} $\uparrow$\hspace{1mm}/ \textcolor{olive}{FPR@95} $\downarrow$\hspace{1mm}}\\
    \hline
    Cora & 90.7/36.8 & 90.7/37.2 & 87.3/50.3 & 89.9/45.7 &
    93.4/29.6 & \textcolor{orange}{\textbf{94.1}}/\textcolor{olive}{\textbf{25.0}}\\
    AmazonCS &  84.1/51.9 & 84.4/51.2 & 81.8/78.8 & 83.6/56.2 &
    91.3/\textcolor{olive}{\textbf{47.2}} & \textcolor{orange}{\textbf{92.3}}/52.0\\
    AmazonPhoto & 94.3/21.7 & 94.3/26.5 & 77.1/59.6 & 94.4/24.1 &
    98.3/7.3 & \textcolor{orange}{\textbf{98.4}}/\textcolor{olive}{\textbf{4.2}}\\
    CoauthorCS & 96.2/19.6 & 96.1/19.8 & 94.0/25.3 & 95.8/22.1 &
    99.1/2.4 & \textcolor{orange}{\textbf{99.6}}/\textcolor{olive}{\textbf{1.4}}\\
    LastFMAsia & 78.5/60.7 & \cellcolor{lightgray}81.1/52.9 & \cellcolor{lightgray}83.4/51.0 & \cellcolor{lightgray}89.6/30.4 &
    \textcolor{orange}{\textbf{91.4}}/\textcolor{olive}{\textbf{25.4}} & 90.5/26.8\\
    Wiki-CS & 80.4/62.5 & 80.4/62.5 & 74.0/74.4 & \cellcolor{lightgray}82.7/54.7 &
    \textcolor{orange}{\textbf{88.7}}/50.0 & 88.6/\textcolor{olive}{\textbf{49.0}}\\
    \bottomrule
  \end{tabular}}
\end{minipage}
\hspace{6mm}
\begin{minipage}{0.35\textwidth}
  \centering
  \caption{Ablation Study}
  \label{tab:ablation}
  \resizebox{\linewidth}{!}{%
  \begin{tabular}{lccc}
    \toprule
    loss & \textcolor{blue}{Acc} $\uparrow$\hspace{1mm} & \textcolor{orange}{AUROC} $\uparrow$\hspace{1mm} & \textcolor{teal}{F1}$\uparrow$\\
    \midrule
    (1) CE & 82.1 & 56.9 & 50.7\\
    (2) CE+con & 86.3 & 90.0 & 78.6\\
    (3) CE+ent & 82.9 & 53.5 & 48.9\\
    (4) CE+dis & 79.3 & 61.2 & 46.1\\
    (5) CE+con+ent & 85.9 & 92.8 & 81.3\\
    (6) CE+con+dis & \textcolor{blue}{\textbf{87.0}} & 89.4 & 78.7\\
    (7) OODGAT & 86.6 & \textcolor{orange}{\textbf{93.1}} & \textcolor{teal}{\textbf{82.2}}\\
    \bottomrule
  \end{tabular}}
\end{minipage}
\end{table*}

\noindent \textbf{Comparison with GKDE.}
We now compare OODGAT with GKDE \cite{uncertainty2020}. To ensure a fair comparison, we test our method on the same datasets used in the original paper and adopt the same preprocessing procedures. (See Appendix \ref{sec:dataset} for details.) We report the AUROC and AUPR for outlier detection in Table \ref{tab:performance_3}, where the results for GKDE are obtained from the original paper. As we can see, although OODGAT is much more efficient than GKDE which requires multiple forward passes due to the Bayesian framework, it still outperforms GKDE in both AUROC and AUPR on all three datasets. The results show that it is not enough to simply embed existing GNNs into the framework of uncertainty computation. Instead, making better use of the information implicit in the graph structure is the key to success.

\subsection{Ablation Study}
The success of OODGAT comes from the combination of the unique propagation strategy and the tailored regularizers for guiding the training process. In this section, we perform ablation analysis in Table \ref{tab:ablation} to demonstrate the importance of each module proposed in Section \ref{sec:reg}. Experiments are done on AmazonComputers using OODGAT-ATT, and the weight for each loss is the same as the best-performing result in Table \ref{tab:performance}. 
In (1), we train the model with only cross-entropy loss. The AUROC for outlier detection is around 50\% which is similar to random guessing, indicating the use of cross-entropy alone is not sufficient to learn the classification of ID and OOD. We then add one of the proposed regularizers in (2),(3) and (4), respectively. The results show that consistency loss can effectively improve the discriminative ability of the binary classifier, while entropy loss and discrepancy loss contribute little or negatively when used without the help of consistency regularizer. This is expected since the other two losses rely on the accurate prediction of the binary classifier, which is learned through consistency loss. Comparing (2) and (5), we find that the entropy loss can further improve the detection ability when used together with consistency loss. Similarly, the comparison between (2) and (6) indicates that the addition of discrepancy regularizer can help the classification of in-distribution samples. The best result is obtained in (7) where we combine all three regularizers with cross-entropy loss. In summary, all regularizers contribute to the final performance, among which consistency loss plays the most important role. For information about the sensitivity of hyperparameters, please see Appendix  \ref{sec:hyper}. 
\begin{table}
  \caption{Comparison with GKDE}
  \label{tab:performance_3}
  \begin{tabular}{l|cc|cc}
    \toprule
    \multirow{2}*{Dataset} & \multicolumn{2}{c|}{AUROC} & \multicolumn{2}{c}{AUPR} \\
    & GKDE & OODGAT & GKDE & OODGAT\\
    \midrule
    Cora & 87.6 & \bftab 91.4 & 78.4 & \bftab 82.9\\
    Citeseer & 84.8 & \bftab 87.7 & 86.8 & \bftab 89.0\\
    Pubmed & 74.6 & \bftab 81.1 & 69.6 & \bftab 76.0\\
    \bottomrule
  \end{tabular}
\end{table}

\section{Conclusion}
In this paper, we propose and study the problem of \textit{graph learning with OOD nodes}. We demonstrate that GNNs are inherently suitable for outlier detection on graphs with high homophily, and propose an end-to-end model OODGAT to tackle the problem of SSOD and SSNC. Extensive experiments show that while existing  methods such as input preprocessing and temperature scaling cannot handle the problem well, OODGAT consistently yields decent performance 
in both in-distribution classification and outlier detection. In the future, we aim to extend OODGAT to more realistic settings such as few-shot learning and incremental learning.

\begin{acks}
This work has been supported by Science and Technology Innovation 2030-Major Project
(2022ZD0208800): Brain Science and Brain-like Research and NSFC General Program
(No.62176215).
\end{acks}

\bibliographystyle{ACM-Reference-Format}
\bibliography{ref}

\appendix
\section{Proof of Proposition}
\label{sec:proof}
We now prove the proposition in Section \ref{sec:analysis}.
\begin{proof}
 The homophily is the fraction of edges in a graph which connect nodes that have the same label. In \cite{pei2020geom}, the node homophily ratio is defined as:

\[h = \frac{1}{|\mathcal{V}|} \sum_{v \in \mathcal{V}} \frac{ | \{ (u,v) : u \in \mathcal{N}(v) \wedge y_u = y_v \} | } { |\mathcal{N}(v)| }\]

Assuming a graph $\mathcal{G}$ whose node homophily ratio w.r.t. $\mathcal{Y}$ is $h$. By definition, we can derive the node homophily ratio w.r.t. $\mathcal{B}$ as:

\[h^{\prime} = \frac{1}{|\mathcal{V}|} \sum_{v \in \mathcal{V}} \frac{ | \mathcal{A}(v) | + \sum_{u \in \mathcal{B}(v) } \delta(f(y_u) = f(y_v)) } { |\mathcal{N}(v)| }\]

where $\mathcal{A}(v)=\{ u : u \in \mathcal{N}(v) \wedge y_u = y_v \}$, $\mathcal{B}(v)=\{u : u \in \mathcal{N}(v) \wedge y_u \neq y_v \}$, and $\mathcal{N}(v) = \mathcal{A}(v) \cup \mathcal{B}(v)$. $f$ is a mapping from $\mathcal{Y}$ to $\mathcal{B}$.

Since for any $v$, we have 
\[|\mathcal{A}(v)| = | \{ (u,v) : u \in \mathcal{N}(v) \wedge y_u = y_v \} |\]
and
\[\sum_{u \in \mathcal{B}(v) } \delta(f(y_u) = f(y_v))\geq0\]
we can derive that
\[
h^{\prime} \geq h
\]
Therefore, if a graph $\mathcal{G}$ is homophilic w.r.t. $\mathcal{Y}$, it is also homophilic w.r.t. to $\mathcal{B}$.
\end{proof}

\section{Joint Classification}
\label{sec:joint}
\begin{figure}[H]
    \centering
    \includegraphics[scale=0.5]{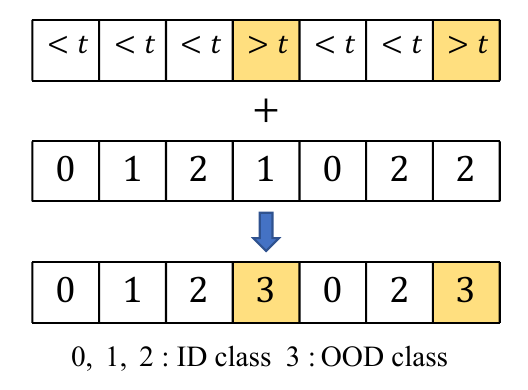}
    \caption{Illustration of joint classification. First row represents the predicted OOD scores, where t is the threshold. Second row represents the predicted labels (3 classes in this case). Combining the two rows, we derive the final result of joint classification (3+1 classes in this case).}
    \label{fig:joint_clf}
\end{figure}

The joint classification includes two stages: first, it classifies nodes to be inliers or outliers according to the OOD scores predicted by the model; then, it assigns in-distribution labels for nodes tagged as ID using their output distributions. An illustration is shown in Figure \ref{fig:joint_clf}. Since the first stage is a binary classification task, the value of weighted-F1 is dependent on the threshold chosen. In the experiments, we report the best F1 value under all possible thresholds.

\section{Dataset Statistics}
\label{sec:dataset}
Statistics for datasets used in the main experiments are listed in Table \ref{tab:dataset}.
\begin{table}[H]
  \caption{Statistics for Main Datasets}
  \label{tab:dataset}
  \begin{tabular}{ccccc}
    \toprule
    Dataset & \#Nodes & \#Edges & \#Features & \#Classes\\
    \midrule
    Cora & 2708 & 10556 & 1433 & 7 \\
    Amazon-Computer & 13752 & 491722 & 767 & 10 \\
    Amazon-Photo & 7650 & 238162 & 745 & 8 \\
    Coauthor-CS & 18333 & 163788 & 6805 & 15 \\
    LastFMAsia & 7624 & 55612 & 128 & 18  \\
    Wiki-CS & 11701 & 297110 & 300 & 10 \\
    \bottomrule
  \end{tabular}
\end{table}

\noindent Experimental setup for main datasets is shown in Table \ref{tab:dataset_setup}.
\begin{table}[H]
  \caption{Experimental Setup for Main Datasets}
  \label{tab:dataset_setup}
  \begin{tabular}{ccc}
    \toprule
    Dataset & OOD class & OOD ratio \\
    \midrule
    Cora & $[0, 1, 3]$ & 0.51 \\
    Amazon-Computer & $[0, 3, 4, 5, 9]$  & 0.49 \\
    Amazon-Photo & $[1, 6, 7]$ & 0.52 \\
    Coauthor-CS & $[0, 1, 2, 3, 4, 9, 13]$ & 0.51 \\
    LastFMAsia & $[1, 2, 3, 4, 5, 9, 10, 12, 17]$ & 0.53 \\
    Wiki-CS & $[0, 2, 4, 5]$ & 0.50 \\
    \bottomrule
  \end{tabular}
\end{table}

\noindent Statistics for citation datasets are listed in Table \ref{tab:citation_dataset}.
\begin{table}[H]
  \caption{Statistics for Citation Datasets}
  \label{tab:citation_dataset}
  \begin{tabular}{ccccc}
    \toprule
    Dataset & \#Nodes & \#Edges & \#Features & \#Classes\\
    \midrule
    Cora & 2708 & 10556 & 1433 & 7 \\
    Citeseer & 3327 & 9104 & 3703 & 6\\
    Pubmed & 19717 & 88648 & 500 & 3\\
    \bottomrule
  \end{tabular}
\end{table}

\noindent Experimental setup for citation datasets is shown in Table \ref{tab:citation_setup}.
\begin{table}[H]
  \caption{Experimental Setup  for Citation Datasets}
  \label{tab:citation_setup}
  \begin{tabular}{ccc}
    \toprule
    Dataset & OOD classes & OOD ratio\\
    \midrule
    Cora & $[0, 1, 2, 3]$ & 0.38\\
    Citeseer & $[0, 1, 2]$ & 0.55 \\
    Pubmed & $[0, 1]$ & 0.40 \\
    \bottomrule
  \end{tabular}
\end{table}

\section{Experiment Details}
\label{sec:exp}
The results from Table \ref{tab:performance} are obtained with the following hyperparameter configurations:
\begin{table}[H]
  \caption{Hyperparameter Configurations of Main Results}
  \label{tab:parameter}
  \resizebox{\linewidth}{!}{%
  \begin{tabular}{cccccccc}
    \toprule
    Data & lr & dropout & $\beta$ & $\gamma$ & $\zeta$ & $\epsilon$ & heads\\
    \midrule
    Cora & 0.01 & 0.5 & 2 & 0.05 & 0.005 & 0.6 & 4\\
    Amazon-Computer & 0.01 & 0.5 & 2 & 0.05 & 0.005 & 0.4 & 4\\
    Amazon-Photo & 0.01 & 0.5 &3 & 0.1 & 0.005 & 0.4 & 4\\
    Coauthor-CS & 0.01 & 0.5 & 4 & 0.05 & 0.005 & 0.6 & 4\\
    LastFMAsia & 0.01 & 0.5 & 3 & 0.3 & 0.005 & 0.5 & 1\\
    Wiki-CS & 0.01 & 0.5 & 3 & 0.2 & 0.005 & 0.5 & 4\\
    \bottomrule
  \end{tabular}}
\end{table}

The results from Table \ref{tab:performance_3} are obtained with hyperparameters in Table \ref{tab:hyper_citation}.
\begin{table}[H]
  \caption{Hyperparameter Configurations of Citation Datasets}
  \label{tab:hyper_citation}
  \resizebox{\linewidth}{!}{%
  \begin{tabular}{cccccccc}
    \toprule
    Data & lr & dropout & $\beta$ & $\gamma$ & $\zeta$ & $\epsilon$ & heads\\
    \midrule
    Cora & 0.01 & 0.5 & 1 & 0.01 & 0.005 & 0.5 & 4\\
    Citeseer & 0.01 & 0.5 & 2 & 0.01 & 0.005 & 0.5 & 4\\
    Pubmed & 0.01 & 0.5 & 1 & 0.1 & 0.005 & 0.4 & 4\\
    \bottomrule
  \end{tabular}}
\end{table}

\section{Influence of Hyperparameters}
\label{sec:hyper}
\begin{figure}
\centering
    \begin{subfigure}{\linewidth}
    \begin{subfigure}[b]{0.49\linewidth}  
        \centering
        \includegraphics[scale=0.3]{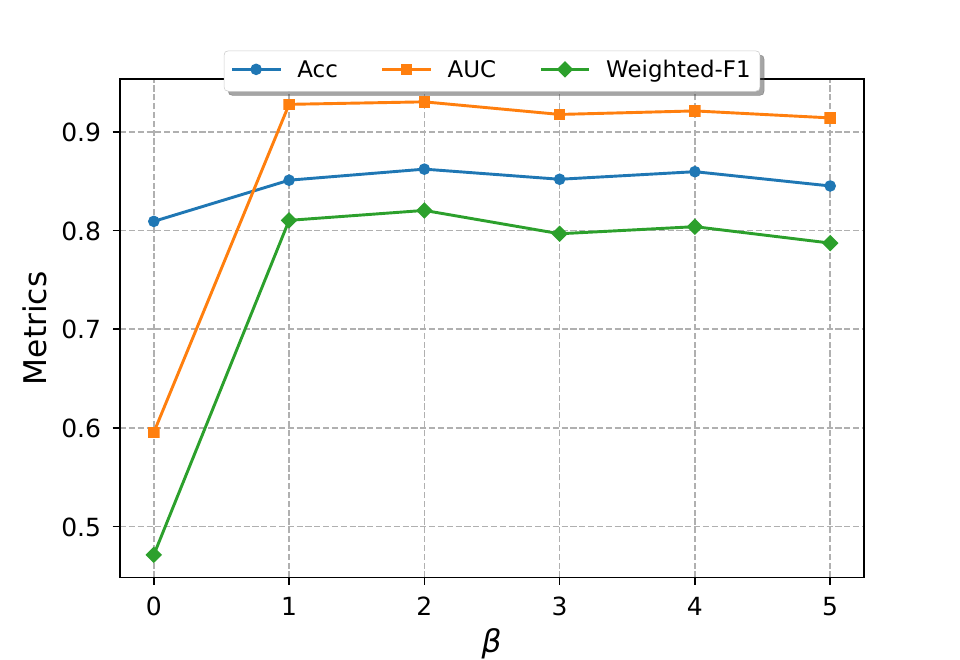}
        \caption{$\beta$: consistency loss}
        \label{fig:sensitivity_a}
    \end{subfigure}
    \hfill
    \begin{subfigure}[b]{0.49\linewidth}  
        \centering
        \includegraphics[scale=0.3]{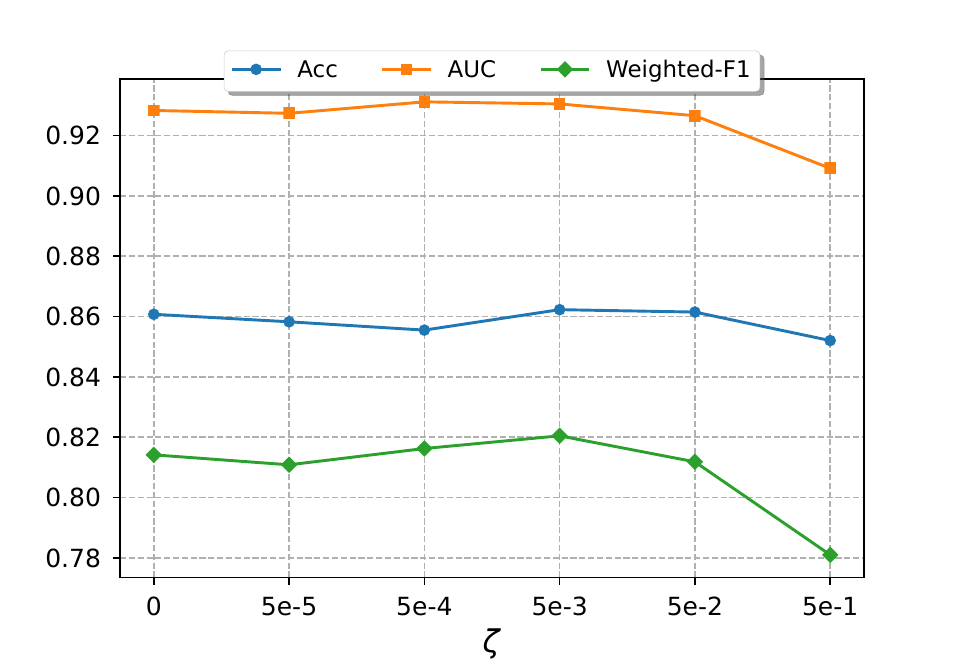}
        \caption{$\zeta$: discrepancy loss}
        \label{fig:sensitivity_b}
    \end{subfigure}
    \end{subfigure}
    
    \begin{subfigure}{\linewidth}
    \begin{subfigure}[b]{0.49\linewidth}  
        \centering
        \includegraphics[scale=0.3]{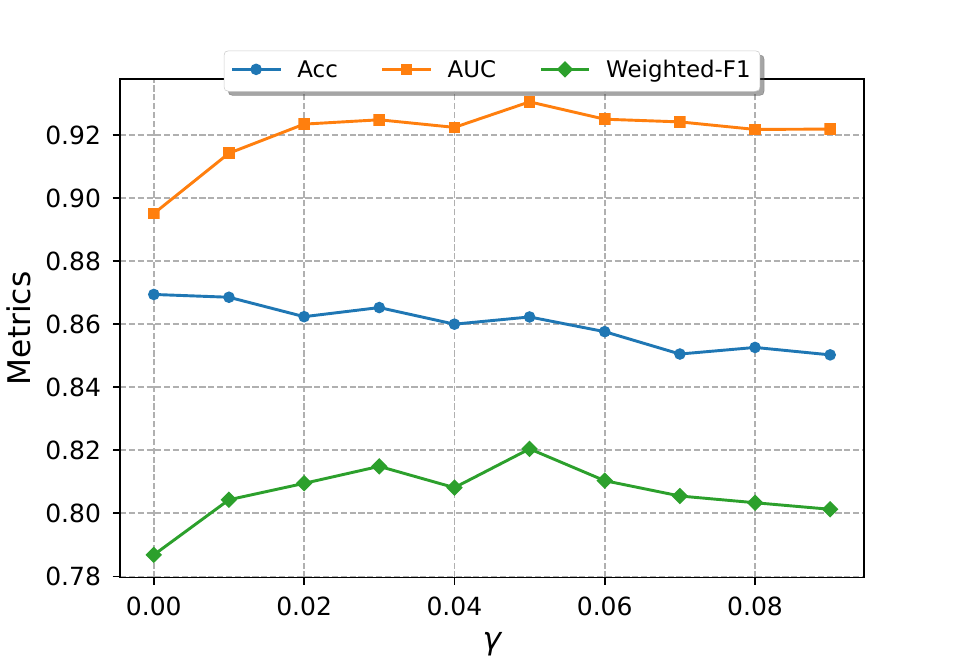}
        \caption{$\gamma$: entropy loss}
        \label{fig:sensitivity_c}
    \end{subfigure}
    \hfill
    \begin{subfigure}[b]{0.49\linewidth}  
        \centering
        \includegraphics[scale=0.3]{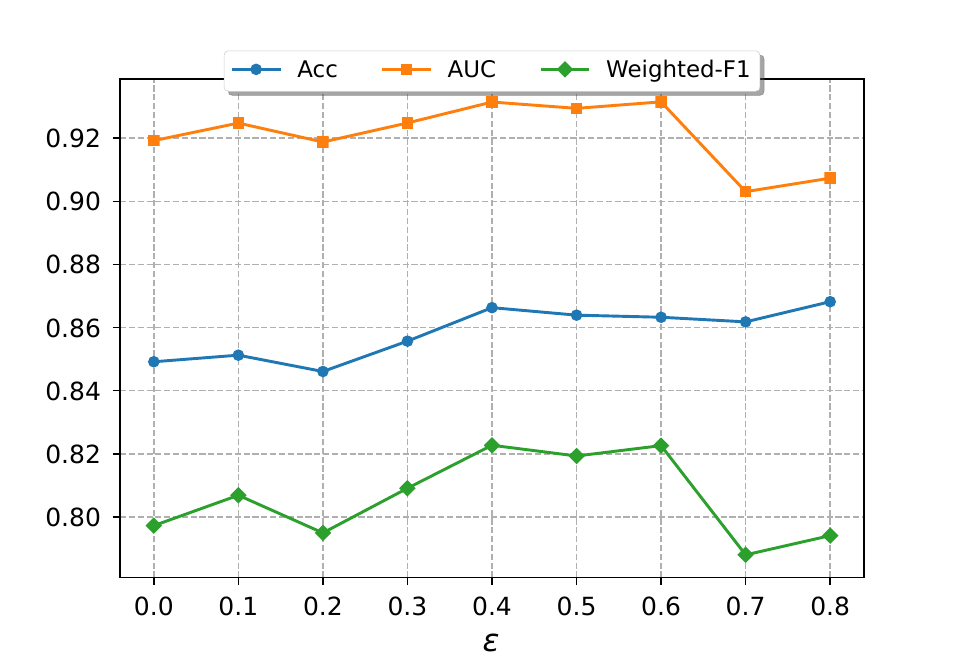}
        \caption{$\epsilon$: threshold}
        \label{fig:sensitivity_d}
    \end{subfigure}
    \end{subfigure}
    \caption{Hyperparameter analysis on AmazonComputers. }
    \Description{}
\end{figure}

The training of OODGAT involves four hyperparameters: $\beta$, $\gamma$, $\zeta$ and $\epsilon$. The former three are the balance parameters of regularizers, while the last is the threshold to determine the set of nodes for which the model encourages uniform distribution. Due to the space limitation, we only present the effect of hyperparameters on AmazonComputers, while similar trends are observed on other datasets. From Figure \ref{fig:sensitivity_a}, we observe that consistency loss is the key to the success of OOD detection. In Figure \ref{fig:sensitivity_b}, the performance is slightly improved when the weight of discrepancy loss reaches around 5e-3. The results in Figure \ref{fig:sensitivity_c} show that while the addition of entropy regularizer can improve detection, it also leads to a decrease in the performance of in-distribution classification. However, by choosing an appropriate trade-off parameter, OODGAT can achieve better detection capability without having too much impact on the classification, thereby improve the overall performance. The effect of the threshold $\epsilon$ is shown in Figure \ref{fig:sensitivity_d}. When the threshold is 0, the entropy loss simply forces all nodes to behave like outliers by increasing the uncertainty of predictions, regardless of their true identity. When moving the threshold to an appropriate range, OODGAT manages to reduce the confidence level of outliers only while leaving the in-distribution data unaffected, resulting in the highest overall performance. 

\section{Visualization}
\label{sec:visual}
\begin{figure}[b]
\centering
    \begin{subfigure}[b]{0.45\linewidth}
        \includegraphics[width=\linewidth]{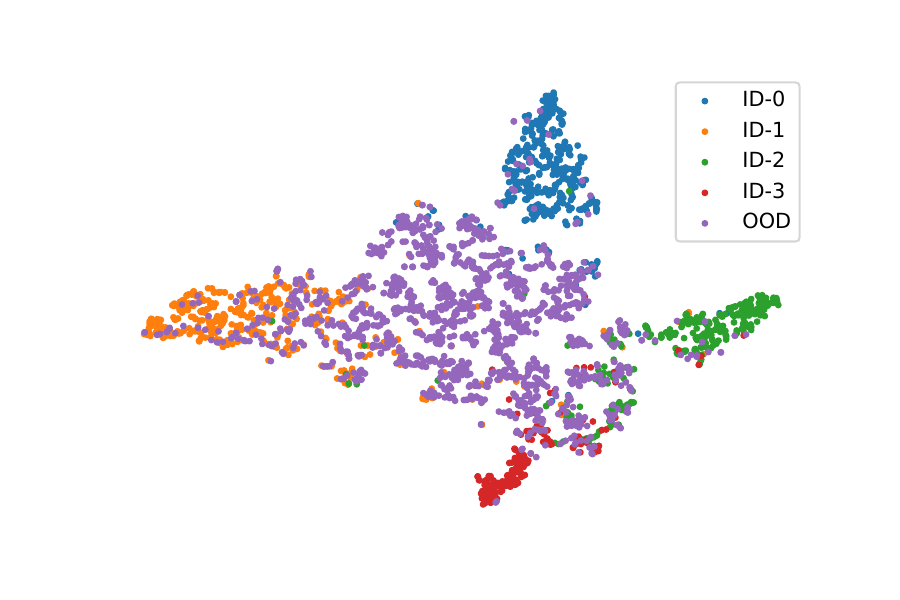}
        \caption{GCN}
    \end{subfigure}
    \hfill
    \begin{subfigure}[b]{0.45\linewidth}
        \includegraphics[width=\linewidth]{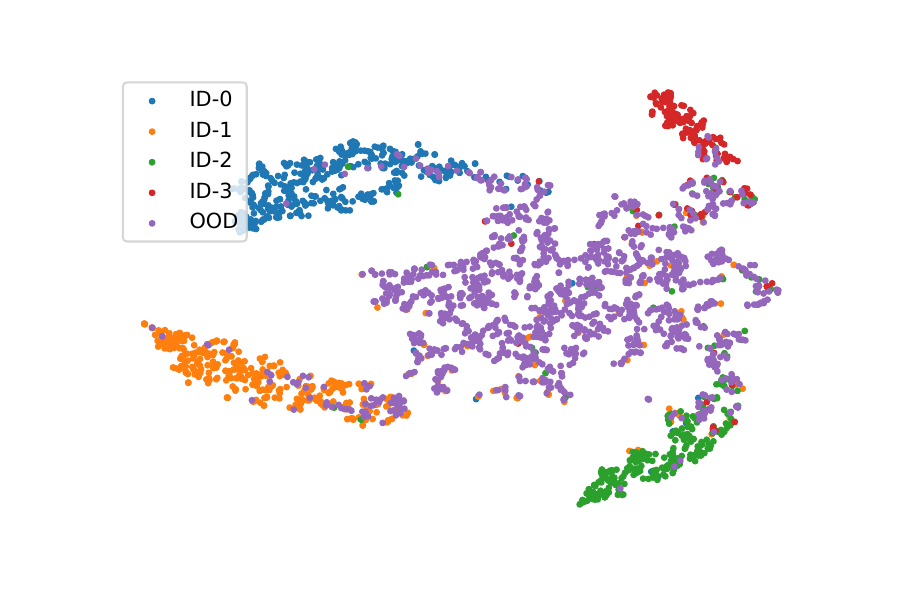}
        \caption{OODGAT}
    \end{subfigure}

    \begin{subfigure}[b]{0.45\linewidth}
            \includegraphics[width=\linewidth]{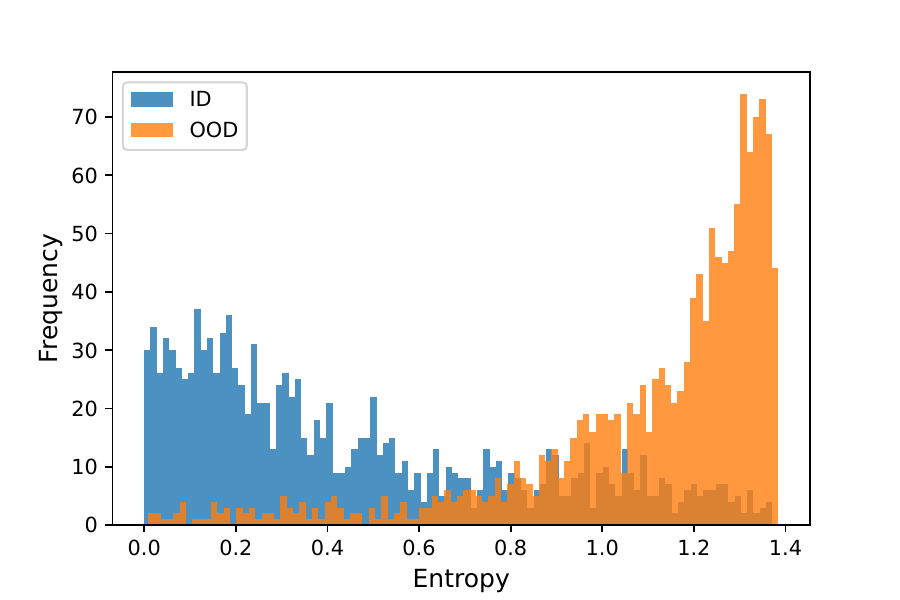}
            \caption{GCN}
    \end{subfigure}
    \hfill
    \begin{subfigure}[b]{0.45\linewidth}
            \includegraphics[width=\linewidth]{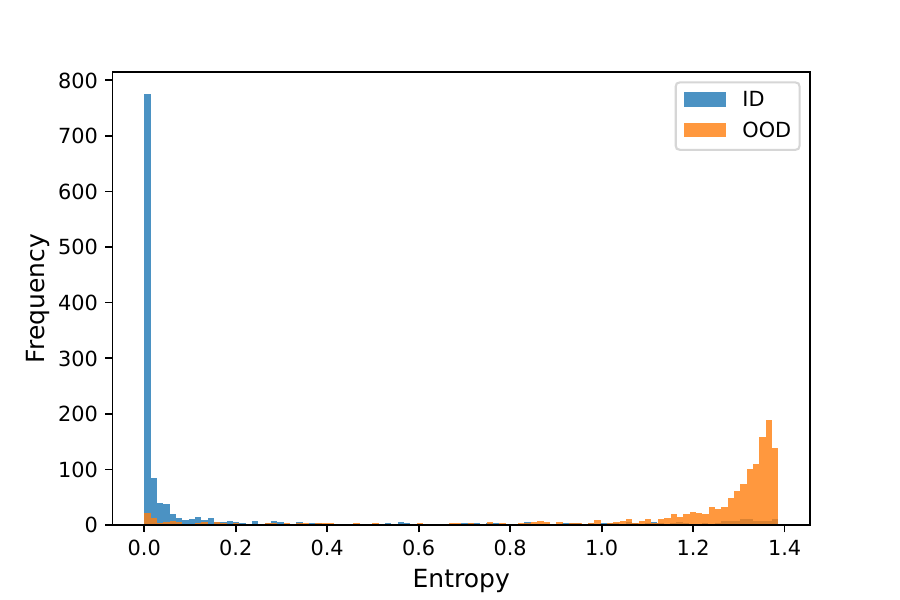}
            \caption{OODGAT}
    \end{subfigure}
    
    \begin{subfigure}[b]{0.45\linewidth}
        \includegraphics[width=\linewidth]{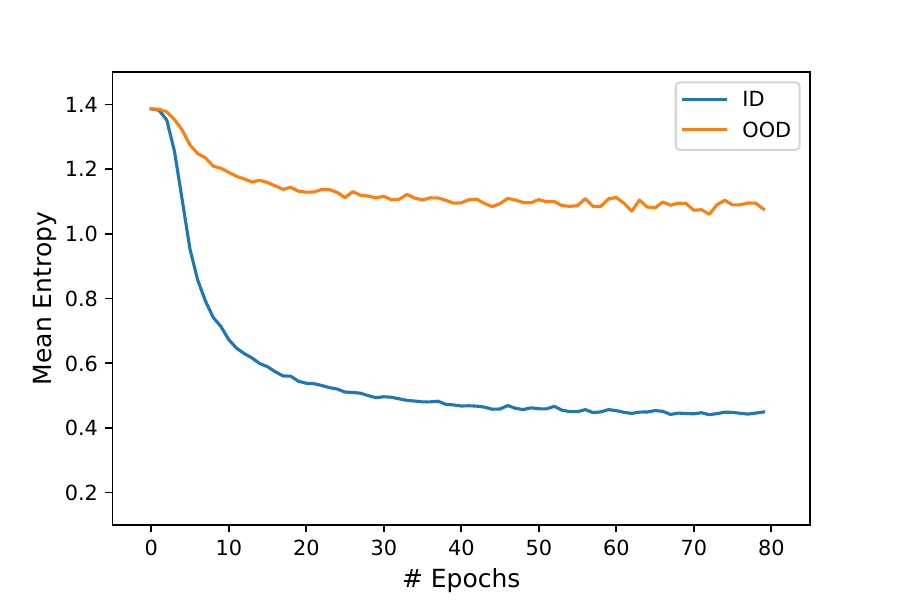}
        \caption{GCN}
    \end{subfigure}
    \hfill
    \begin{subfigure}[b]{0.45\linewidth}
        \includegraphics[width=\linewidth]{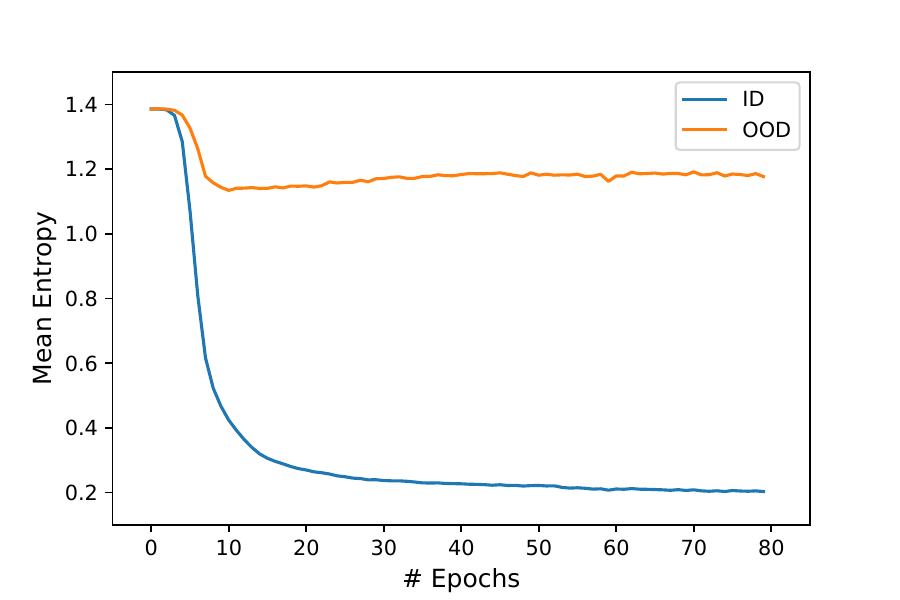}
        \caption{OODGAT}
    \end{subfigure}
    \caption{Visualization of GCN and OODGAT. Experiments done on Cora. (a) and (b): t-SNE plot of the latent space, (c) and (d): distribution of nodes' predictive uncertainties, (e) and (f): training dynamics of the mean entropy of inliers and outliers. In (a) and (b), OODGAT shows a clearer boundary between ID and OOD classes. In (c) and (d), OODGAT produces scores with less overlap between ID and OOD. In (e) and (f), OODGAT maintains a larger gap between the entropy of inliers and outliers throughout the training phase.}
    \label{fig:visualize}
\end{figure}

We present some visualizations about GCN and OODGAT in Figure \ref{fig:visualize}. 

\end{document}